\documentclass[runningheads]{llncs}

\usepackage[T1]{fontenc}
\usepackage{graphicx}
\usepackage{booktabs}
\usepackage{amssymb} 
\usepackage{tikz}
\usepackage{hyperref}
\usetikzlibrary{positioning, calc, arrows.meta, matrix, fit}
\newcommand{\corr}{\textsuperscript{*}}

\begin{document}

\title{L-GTA: Latent Generative Modeling for Time Series Augmentation}
\titlerunning{Latent Generative Modeling for Time Series Augmentation}

\author{Luis~Roque~\corr\inst{1} \and Vitor~Cerqueira\inst{3} \and Carlos~Soares\inst{1,2} \and Luis~Torgo\inst{4}}

\authorrunning{Roque et al.}
\toctitle{L-GTA: Latent Generative Modeling for Time Series Augmentation}
\tocauthor{Luis Roque, Vitor Cerqueira, Carlos Soares, Luis Torgo}

\institute{Faculdade de Engenharia da Universidade do Porto, Porto, Portugal \\
\email{luis\_roque@live.com} \and
Fraunhofer Portugal AICOS, Portugal \and
University of Coimbra, Coimbra, Portugal \and
Dalhousie University, Halifax, Canada
}

\maketitle              

\begin{abstract}
Data augmentation is becoming increasingly important across various areas of time series analysis, including forecasting, classification, and anomaly detection. We introduce the Latent Generative Temporal Augmentation (L-GTA) model, a generative approach based on a Variational Autoencoder with a Bi-LSTM backbone and temporal self-attention. The model learns a latent representation for each timestep and applies controlled perturbations such as jittering, magnitude warping, or drift. We define an equivariance objective to further encourage consistency between latent space and data space transformations. As a result, the augmented samples show predictable and interpretable transformation signatures. We evaluate L-GTA on several real-world datasets against SOTA generative methods, including TimeGAN, TimeVAE, and Diffusion-TS, as well as direct transformation approaches. Across experiments on downstream forecasting, distribution fidelity, and controllability of transformation intensity, L-GTA consistently outperforms competing approaches. In downstream forecasting, it reduces prediction error by up to 26\% compared to the strongest generative method and 27\% relative to using the original data without augmentation.

\end{abstract}

\section{Introduction}

In the era of big data, time series analysis has emerged as a critical tool in various domains ranging from financial markets~\cite{CHENG2022108218}, climate forecasting~\cite{Fathi2022} to retail sales predictions~\cite{KARMY201959}. Accurate and insightful time series analysis can lead to predictive models that provide crucial information for decision-making processes. Nevertheless, the performance of these models significantly depends on the quality and quantity of the available data. Also, collecting labeled time series data for a specific task or domain can be challenging.

Similarly, real-world time series data often exhibit complex dependencies, non-linear dynamics, and high dimensionality. They may also be affected by noise, irregular sampling, and missing values, further complicating their analysis and modeling. As a result, models trained on such data often fail to generalize well, leading to poor predictive performance when applied to unseen data. In such scenarios, data augmentation techniques, which generate synthetic data samples from the original data, offer a promising solution. They increase the quantity of the data and improve model robustness by providing diversified data instances~\cite{Wen_2021}.
The problem with traditional data augmentation techniques for time series data, such as jittering, scaling, and warping, is that they are relatively simple and may not adequately capture the complexities often found in relevant datasets. Furthermore, these techniques are difficult to apply in a controlled manner, which may introduce artificial distortions that deviate significantly from real-world scenarios. Consequently, this reduces the practical utility of the models trained on such data.

Recent advances in generative modeling have created new opportunities for time series augmentation. Methods based on Generative Adversarial Networks (GANs), Variational Autoencoders (VAEs), and diffusion models learn the data distribution and generate synthetic samples from the learned representation. While these approaches can produce realistic time series, they present practical challenges. GANs are often difficult to train due to instability and mode collapse. Diffusion models typically require substantial computational resources because generation involves iterative denoising from random noise. More broadly, these methods rely on stochastic generation, which limits control over the characteristics of the generated samples and makes targeted augmentation difficult.

To address these limitations, we introduce the \textit{Latent Generative Temporal Augmentation} (L-GTA) model (Figure~\ref{fig:lgta_architecture}). L-GTA is a generative model designed to produce controllable and interpretable augmentations of time series data. The model is based on a VAE architecture with a Bi-directional Long Short-Term Memory (Bi-LSTM) backbone and a temporal multi-head self-attention mechanism. This architecture enables the model to capture both short-term temporal dependencies and longer-range relationships within the data.


\begin{figure*}[t]
\centering
\resizebox{0.9\textwidth}{!}{%
\begin{tikzpicture}[
    >=Stealth,
    encoder_col/.style  = {fill=blue!6,   draw=blue!45!black,  line width=0.8pt},
    decoder_col/.style  = {fill=orange!6, draw=orange!50!black, line width=0.8pt},
    latent_col/.style   = {fill=green!6,  draw=green!45!black,  line width=0.8pt},
    transf_col/.style   = {fill=violet!6, draw=violet!45!black, line width=0.8pt},
    io_col/.style       = {fill=gray!6,   draw=gray!55,         line width=0.8pt},
    loss_col/.style     = {fill=red!5,    draw=red!45!black,    line width=0.8pt},
    equiv_col/.style    = {fill=teal!5,   draw=teal!50!black,   line width=0.8pt,
                           dashed, dash pattern=on 3pt off 2pt},
    ctx_col/.style      = {fill=yellow!6, draw=yellow!50!black, line width=0.8pt},
    bigblock/.style = {
        rectangle, rounded corners=4pt,
        minimum height=3.4cm, minimum width=4.2cm,
        align=center},
    medblock/.style = {
        rectangle, rounded corners=4pt,
        minimum height=3.4cm, minimum width=3.6cm,
        align=center},
    innerblock/.style = {
        rectangle, rounded corners=2pt,
        minimum height=0.6cm, minimum width=3.0cm,
        align=center, font=\footnotesize, line width=0.5pt},
    ioblock/.style = {
        rectangle, rounded corners=3pt,
        minimum height=1.2cm, minimum width=1.8cm,
        align=center, font=\normalsize, io_col},
    prepblock/.style = {
        rectangle, rounded corners=3pt,
        minimum height=1.2cm, minimum width=2.4cm,
        align=center, font=\footnotesize, io_col},
    arrow/.style = {->, line width=1.2pt, draw=gray!60},
    thinarrow/.style = {->, line width=0.8pt, draw=gray!50},
    dasharrow/.style = {->, line width=0.8pt, draw=teal!50!black,
                        dashed, dash pattern=on 3pt off 2pt},
    dimlabel/.style = {font=\footnotesize\itshape, text=gray!65!black},
    seclabel/.style = {font=\normalsize\bfseries},
]


\node[ioblock] (input) at (0, 0)
    {Input\\[2pt]$z \in \mathbb{R}^{T \times S}$};

\node[bigblock, encoder_col] (enc) at (5.4, 0) {};
\node[seclabel, text=blue!45!black, anchor=north]
    at ([yshift=-4pt]enc.north) {Encoder\; $\mathcal{E}$};
\node[innerblock, encoder_col] (enc_bilstm)
    at ([yshift=0.05cm]enc.center)
    {Bi-LSTM $\to$ LayerNorm};
\node[innerblock, encoder_col, below=0.3cm of enc_bilstm] (enc_attn)
    {Temporal Self-Attention\\[-1pt]
     {\scriptsize pos.\ embed \textbullet{} MHA \textbullet{} FFN}};

\node[medblock, latent_col, minimum height=3.8cm] (lat) at (10.4, 0) {};
\node[seclabel, text=green!38!black, anchor=north]
    at ([yshift=-4pt]lat.north) {Latent Space};
\node[innerblock, latent_col, minimum width=2.4cm]
    (lat_mu) at ([yshift=0.1cm]lat.center)
    {$\mu_t,\;\log\Sigma_t$};
\node[innerblock, latent_col, minimum width=2.4cm,
      below=0.25cm of lat_mu] (lat_sample)
    {Sampling\\[-1pt]{\scriptsize reparam.\ trick}};
\node[dimlabel, below=0.1cm of lat_sample]
    {$v_t \sim \mathcal{N}(\mu_t,\, \Sigma_t)$};


\node[medblock, transf_col, minimum height=4.2cm, minimum width=4.2cm] (transf) at (10.4, -5.2) {};
\node[seclabel, text=violet!45!black, anchor=north]
    at ([yshift=-4pt]transf.north) {Latent Transforms};
\node[innerblock, transf_col, minimum width=2.4cm]
    (transf_fn) at ([yshift=0.45cm]transf.center)
    {$v' = \mathcal{T}_\eta(v,\;\eta)$};
\node[font=\footnotesize, text=violet!40!black, align=center,
      below=0.25cm of transf_fn] (transf_list)
    {jitter \textbullet{} scaling \textbullet{} mag.\ warp\\
     drift \textbullet{} trend};
\node[dimlabel, align=center, below=0.25cm of transf_list]
    {chainable: $\mathcal{T}_n \!\circ\! \cdots \!\circ\! \mathcal{T}_1$};


\node[bigblock, decoder_col, minimum height=4.2cm] (dec) at (5.4, -5.2) {};
\node[seclabel, text=orange!50!black, anchor=north]
    at ([yshift=-4pt]dec.north) {Decoder\; $\mathcal{D}$};
\node[font=\footnotesize\itshape, text=orange!45!black,
      anchor=north] at ([yshift=-0.5cm]dec.north)
    {$+\; z_{\mathrm{proj}}(v')$ skip connection};
\node[innerblock, decoder_col] (dec_bilstm)
    at ([yshift=-0.05cm]dec.center) {Bi-LSTM};
\node[innerblock, decoder_col, below=0.3cm of dec_bilstm] (dec_fc)
    {FC };

\node[ioblock] (output) at (0, -5.2)
    {Output\\[2pt]$\tilde{z} \in \mathbb{R}^{T \times S}$};



\draw[arrow] (input) -- (enc);
\draw[arrow] (enc)   -- (lat);

\draw[thinarrow] (enc_bilstm) -- (enc_attn);

\draw[arrow] (lat.south) -- (transf.north);

\draw[arrow] (transf.west) -- (dec.east);
\draw[arrow] (dec) -- (output);



\node[rectangle, rounded corners=4pt, loss_col,
      minimum height=1.1cm,
      align=center, font=\small, inner sep=8pt]
    (loss) at (4.8, -8.5)
    {$\mathcal{L}
       = \mathcal{L}_{\mathrm{recon}}
       + \beta\,\mathcal{L}_{\mathrm{KL}}
       + \lambda_{\mathrm{equiv}}\,\mathcal{L}_{\mathrm{equiv}}$};
\node[dimlabel, below=3pt of loss, align=center]
    {$\beta$: KL annealing
     \qquad $\lambda_{\mathrm{equiv}}$: equivariance weight};

\draw[->, line width=0.9pt, draw=red!40!black]
    (dec.south) -- (dec.south |- loss.north);

\node[rectangle, rounded corners=4pt, equiv_col,
      minimum height=1.65cm, minimum width=5cm,
      align=center] (equiv) at (10.4, -8.5) {};
\node[seclabel, text=teal!45!black, anchor=north]
    at ([yshift=-4pt]equiv.north) {Equivariance};
\node[font=\footnotesize, align=center] at ([yshift=-0.1cm]equiv.center)
    {$\mathcal{D}\bigl(\mathcal{T}(v); c\bigr)
      \approx
      \mathcal{T}\bigl(\mathcal{D}(v; c)\bigr)$\\[2pt]
     {\scriptsize\itshape decode$\,\circ\,$transform
      $\approx$
      transform$\,\circ\,$decode}};

\draw[dasharrow] (equiv.west) -- (loss.east);

\end{tikzpicture}
}

\caption{
Architecture of L-GTA. The {Encoder} $\mathcal{E}$ processes each window with a Bi-LSTM and temporal multi-head self-attention with learned positional embeddings, producing per-timestep Gaussian parameters $(\mu_t, \Sigma_t)$. Latent vectors $v_t$ are sampled via the reparameterization trick. Parametric {latent-space transformations} $\mathcal{T}_\eta$ are applied to $v$ and can be composed arbitrarily. The {Decoder} $\mathcal{D}$ reconstructs the series from the transformed latent $v'$ using a Bi-LSTM with a learned skip connection ($z_{\mathrm{proj}}$). An {equivariance} loss $\mathcal{L}_{\mathrm{equiv}}$ encourages $\mathcal{D}(\mathcal{T}(v)) \approx \mathcal{T}(\mathcal{D}(v))$.
}
\label{fig:lgta_architecture}
\end{figure*}

One key component of L-GTA is that it learns a latent representation for each timestep of the time series. Instead of generating new samples by random latent sampling, we apply parametric transformations directly to the latent trajectory. We used different transformations, such as jittering, scaling, magnitude warping, drift, and additive trends. By manipulating the latent representation rather than the observed series, the decoder propagates the transformations in a coherent manner that preserves the statistical structure of the original data.

To further improve the reliability of the generated samples, L-GTA incorporates an equivariance regularization objective. It encourages consistency between transformations applied in latent space and their corresponding effects in observation space. In other words, applying a transformation before decoding should produce a similar result to applying the corresponding transformation directly to the decoded series. This constraint promotes predictable transformation behavior and improves the controllability of the augmentation process.

We evaluate L-GTA on several real-world time-series datasets and compare it with both direct transformation methods and SOTA generative models, including TimeGAN~\cite{timegan}, TimeVAE~\cite{desai2021timevaevariationalautoencodermultivariate}, and Diffusion-TS~\cite{yuan2024diffusiontsinterpretablediffusiongeneral}. The evaluation considers complementary perspectives: distributional fidelity, reconstruction quality, downstream forecasting performance, and the controllability of latent transformations. 

Across all evaluations, L-GTA consistently outperforms the competing approaches. In downstream forecasting, it reduces the average error by close to {26\%} compared to the strongest generative benchmark and by about {27\%} relative to using the original data without augmentation. The improvement is even larger compared to direct transformations (about {33\%}) and diffusion- or GAN-based methods, where the reduction exceeds {40--50\%}. These gains are achieved while preserving the statistical structure of the original series and maintaining high controllability of the transformations, demonstrating that L-GTA provides a substantially more effective approach to time series augmentation than existing generative or direct methods.

The primary contributions of this paper are as follows:

\begin{itemize}

\item We propose L-GTA, a generative time series augmentation model based on a CVAE with a Bi-LSTM backbone and temporal self-attention that learns a latent representation at each timestep. The model enables controlled augmentation by applying parametric transformations directly to latent trajectories, allowing predictable and interpretable variations in the generated time series.

\item To the best of our knowledge, this is the first work to introduce an equivariance regularization objective for controllable time series augmentation. This objective enforces consistency between transformations applied in latent space and their corresponding effects in observation space, improving the reliability and controllability of the generated samples.

\item We conduct an extensive empirical evaluation comparing L-GTA with direct transformation methods and SOTA generative models, demonstrating improved fidelity, controllability, and downstream forecasting performance.

\end{itemize}

All experiments are fully reproducible, and the methods and time series data are available as a publicly available code repository.\protect\footnotemark{}

\footnotetext{\href{https://github.com/luisroque/latent-generative-modeling-time-series-augmentation}{Code repository}}

\section{Notation and Background}
\label{chap:background}

\subsection{Time Series}

Let us consider a set of $S$ related univariate time series, represented as 
$\mathcal{Z} = \{z^i_t : t = 1, \dots, T,\; i = 1, \dots, S\}$, where 
$z^i_{1:T} = [z^i_1, z^i_2, \ldots, z^i_T]$ denotes the observed values of the 
$i$-th time series up to the final observation time $T$. Each observation 
$z^i_t \in \mathbb{R}$ indicates the value of series $i$ at time $t$. For 
convenience, we denote $\mathbf{z}^i = z^i_{1:T}$ as the complete observed 
time series for the $i$-th series.

Time series augmentation aims to generate additional series derived from the 
original data. We denote the augmented dataset as 
$\tilde{\mathcal{Z}} = \{\tilde{z}^i_t : t = 1, \dots, T,\; i = 1, \dots, S\}$, 
where $\tilde{\mathbf{z}}^i = \tilde{z}^i_{1:T}$ represents a synthetic series 
generated from the original series $\mathbf{z}^i$. The goal is to expand the 
dataset for downstream analysis or model training while preserving the 
statistical properties of the original time series.

\subsection{Bi-directional Long Short-Term Memory Networks}

Long Short-Term Memory (LSTM) networks are a type of Recurrent Neural Network (RNN) that effectively handle the issue of long-term dependencies in sequences. They accomplish this with a gating mechanism that selectively forgets and updates the cell state at each time step.

Bidirectional LSTM (Bi-LSTM) networks \cite{bilstm} are an extension of LSTMs that process the data in both forward and backward directions. Given an input sequence $\mathbf{z} = (z_1, z_2, ..., z_T)$, a Bi-LSTM consists of a forward $\overrightarrow{\mathrm{LSTM}}$ and a backward $\overleftarrow{\mathrm{LSTM}}$. The forward LSTM reads the sequence in the forward direction to produce a sequence of hidden states $\overrightarrow{h_t}$, and the backward LSTM reads the sequence in the backward direction to produce a sequence of hidden states $\overleftarrow{h_t}$. For a time step $t$, the forward and backward hidden states are given by $\overrightarrow{h_t} = \overrightarrow{\mathrm{LSTM}}(z_t, \overrightarrow{h_{t-1}})$ and $\overleftarrow{h_t} = \overleftarrow{\mathrm{LSTM}}(z_t, \overleftarrow{h_{t+1}})$.

The hidden states of the Bi-LSTM at each time step $t$, $h_t$, are then obtained by concatenating the forward and backward hidden states $h_t = [\overrightarrow{h_t}; \overleftarrow{h_t}]$. Bi-LSTMs capture both past and future information for a given time step.

\subsection{Variational Autoencoders}

Autoencoders (AEs) are neural network architectures that aim to reconstruct their input by first encoding it into a latent space and then decoding it back to the original space. Formally, an autoencoder consists of two components: an encoder function $\phi$ and a decoder function $\psi$. Given an input $z \in \mathbb{R}^d$, the encoder maps $z$ to a latent representation, $v = \phi(z), v \in \mathbb{R}^p$. The decoder then maps $v$ back to the original space to produce a reconstruction $\tilde{z} = \psi(v), \tilde{z} \in \mathbb{R}^d$. Autoencoders are typically trained by minimizing the reconstruction error between the original input $z$ and the reconstruction $\tilde{z}$.

Variational Autoencoders (VAEs) \cite{kingma2022autoencoding} are a generative variant of autoencoders that introduce a probabilistic approach to the encoding process. Instead of directly mapping an input $z$ to a deterministic latent representation $v$, a VAE maps $z$ to a distribution over the latent space. The encoder of a VAE, also known as the recognition model, thus outputs the parameters of a Gaussian distribution, typically the mean $\mu$ and the diagonal covariance $\Sigma$. Given an input $z$, the encoder outputs $\left(\mu, \Sigma\right) = \phi(z)$. A latent variable $v$ is then sampled from this Gaussian distribution, $v \sim \mathcal{N}(\mu, \Sigma)$. The decoder of a VAE maps $v$ back to the original space to produce $\tilde{z}$, a reconstruction of $z$ (a process similar to AEs). The VAE is trained by maximizing the Evidence Lower Bound (ELBO) on the log-likelihood of the data:

\begin{equation}
\log p(z) \geq \mathbb{E}{q(v|z)} [\log p(z|v)] - D{KL}(q(v|z) || p(v))
\end{equation}
where $q(v|z)$ is the approximate posterior, $p(v)$ is the prior, and $D_{KL}$ is the Kullback-Leibler divergence.

\subsection{Attention for Recurrent Temporal Representations}

Self-attention mechanisms have become a standard tool for modeling dependencies in sequential data, including time series (e.g.,~\cite{nie2023timeseriesworth64}). Given a sequence of representations, attention computes similarity scores between elements of the sequence and uses them to construct context vectors as weighted combinations of the representations. Multi-head attention extends this idea by applying several attention operations in parallel with different learned projections, enabling the model to capture different relationships within the sequence.

Attention mechanisms can also be combined with recurrent encoders for time-series forecasting~\cite{AIKP2023LSTMmultiattn}. Let ${h_t} = (h_1, \ldots, h_T)$ denote the hidden states produced by a recurrent encoder such as a Bi-LSTM. {Temporal attention} assigns an importance weight to each hidden state in order to determine which past time steps are most relevant for prediction. The resulting context vector is obtained as a weighted aggregation of the encoder states,

\begin{equation}
c_t = \sum_{j=1}^{T} a_{tj} h_j,
\end{equation}
where $a_{tj}$ denotes normalized attention weights computed from similarity scores between hidden states. In addition to temporal attention, {feature attention} can be used to determine the relative importance of the input variables. For an observation $z_t \in \mathbb{R}^d$, feature attention assigns a weight vector $\alpha_t \in \mathbb{R}^d$ and produces a reweighted input representation

\begin{equation}
\tilde{z}_t = \alpha_t \odot z_t,
\end{equation}
where $\odot$ denotes element-wise multiplication. By combining temporal and feature attention, the model can selectively focus on both informative time steps and relevant variables.

Attention mechanisms can also be incorporated into the variational framework of VAEs. Instead of directly using the deterministic attention context vector $c_t$, the attention-derived representation can be modeled as a latent stochastic variable. In this formulation, a latent variable $v_t$ is inferred from the attention representation through an approximate posterior

\begin{equation}
q(v_t \mid c_t) = \mathcal{N}(\mu_t, \Sigma_t),
\end{equation}
where the parameters $\mu_t$ and $\Sigma_t$ are learned functions of the attention-derived context vector. Sampling follows the reparameterization trick~\cite{kingma2022autoencoding}. The stochastic variable $v_t$ replaces the deterministic context representation in the generative model, enabling the attention mechanism to capture uncertainty in the aggregated temporal information.

\subsection{Data Augmentation for Time Series}
\label{chap:related_work}

Data augmentation aims to generate synthetic samples that expand the input space while preserving the statistical structure and labels of the original dataset, reducing the need for additional real-world data collection~\cite{Wen_2021}. While augmentation has been widely successful in domains such as computer vision, applying it to time series is more challenging due to temporal dependencies, non-linear dynamics, and task-specific constraints~\cite{Wen_2021}. Existing approaches can be broadly grouped into several categories. Transformation-based methods, including jittering, scaling, and magnitude warping~\cite{Wen_2021}, are simple and widely used for tasks such as time series classification but may fail to capture complex temporal patterns and can introduce unrealistic distortions. Pattern mixing approaches combine multiple time series to generate new samples but may overlook global temporal dependencies and risk overfitting due to repeated patterns~\cite{Iwana_2021}. Decomposition-based methods generate new data from extracted components such as trend or seasonal patterns~\cite{decomp_aug}, although they typically provide limited diversity. Autoregressive models have also been used to generate synthetic sequences by leveraging past observations to predict future values~\cite{gratis}, but they are generally more suitable for linear and stationary processes. 

More recently, generative models have gained attention for time series augmentation as they attempt to learn the underlying data distribution and generate realistic synthetic samples~\cite{Wen_2021}. GAN-based approaches such as TimeGAN~\cite{timegan} and SigWGAN~\cite{ni2021sigwasserstein} adapt adversarial learning to sequential data by combining generative modeling with temporal representation learning. However, GANs are often difficult to train due to unstable optimization and issues such as mode collapse. Variational approaches, including VAEs and Conditional VAEs (CVAEs), have also been explored for time series augmentation~\cite{Wen_2021,vae_aug}, frequently incorporating recurrent architectures inspired by variational RNNs~\cite{vrnn} to capture temporal dependencies. More recently, diffusion-based models such as Diffusion-TS~\cite{yuan2024diffusiontsinterpretablediffusiongeneral} have been proposed for time series generation, leveraging iterative denoising processes to produce high-quality samples. Despite their effectiveness, these generative approaches typically rely on stochastic generation, either through latent variable sampling (GANs and VAEs) or iterative noise-to-data transformations (diffusion models), which can make it difficult to control the characteristics of the generated time series.

\section{L-GTA}
\label{chap:lt_gam}

We introduce L-GTA, a method for generating semi-synthetic time series data that integrates temporal self-attention, BiLSTMs, a VAE, an equivariance objective, and traditional time series augmentation techniques. This architecture enables the efficient generation of diverse and controllable time series while preserving the statistical properties of the original data.

\subsection{Generative Model}
\label{chap:gen_model_architecture}

L-GTA is a generative model that combines Bi-LSTMs and a VAE with a temporal multi-head self-attention mechanism to generate semi-synthetic time series data. The self-attention layer augments the Bi-LSTM encoder to capture long-range temporal dependencies while preserving a latent representation at each timestep, enabling transformations to be applied directly in latent space.

The L-GTA model processes an input time series $z$ and feeds it into a Bi-LSTM encoder. A temporal multi-head self-attention block is then applied to the sequence of Bi-LSTM hidden states, enabling the model to attend to different parts of the sequence and capture long-term dynamics. This process can be formalized as follows:
\begin{equation}
\left(\mu_t, \Sigma_t\right) = \phi(\mathrm{SA}(h_t)),
\end{equation}
where $h_t$ represents the Bi-LSTM hidden state at time $t$, and $\mathrm{SA}(h_t)$ denotes the application of a temporal multi-head self-attention mechanism over the sequence of hidden states, enriching their representation. The function $\phi$ then maps this enriched representation to the parameters $\mu_t$ and $\Sigma_t$ of a Gaussian distribution, from which the latent space representation $v_t$ is sampled:
\begin{equation}
v_t \sim \mathcal{N}(\mu_t, \Sigma_t).
\end{equation}

This enables the production of various plausible time series variations. The bidirectional structure of the Bi-LSTM, augmented with self-attention, is particularly adept at learning complex temporal structures in the data. This encoder produces a temporal latent space $v = (v_1, \dots, v_T)$.

The decoding process reconstructs the time series $\tilde{z}$ from the latent trajectory $v$. We denote this operation by the decoder
\begin{equation}
\tilde{z} = \mathcal{D}(v),
\end{equation}
which maps a latent sequence $v$ to a reconstructed series. In practice, $\mathcal{D}$ is implemented using a Bi-LSTM decoder:
\begin{equation}
\tilde{y}_t = \mathrm{Bi\mbox{-}LSTM}(v_t, \tilde{h}_{t-1}), \quad
\tilde{z}_t = \psi(\tilde{y}_t),
\end{equation}
where $\tilde{y}_t$ denotes the decoder output at time $t$, and $\psi$ maps it to the reconstructed observation $\tilde{z}_t$.

A distinctive feature of L-GTA is its ability to generate datasets with controlled variability, enabling the systematic variation of the similarity between the original and generated data. The process involves applying parametric transformations to the latent space of the model, which can be defined as $v^{\prime i} = \mathcal{T}_{\eta}(v^{i}, \eta)$. Here, $\mathcal{T}$ represents the transformation function, and $\eta$ its parameters. The transformed latent representation $v^{\prime i}$ is then decoded to yield the transformed time series $\tilde{z}^{\prime i} = \mathcal{D}(v^{\prime i})$, maintaining consistency with the statistical properties of the original series.

Furthermore, L-GTA supports the sequential chaining of multiple transformations in the latent space, thus enabling the generation of an extensive and diverse set of semi-synthetic time series. This is formulated as:
\begin{equation}
v^{\prime i} = \mathcal{T}_n\bigl( \dots \mathcal{T}_2(\mathcal{T}_1(v^{i}, \eta_1), \eta_2) \dots, \eta_n \bigr),
\end{equation}
where $\{\mathcal{T}_1, \mathcal{T}_2, ..., \mathcal{T}_n\}$ are the transformation functions, applied in sequence with their respective parameters $\{\eta_1, \eta_2, ..., \eta_n\}$. Each transformation in the chain is applied to a normalized version of the latent trajectory, and the result is re-normalized before applying the next transformation. This ensures that non-additive transformations (such as multiplicative or spline-based warps) remain well-behaved across different magnitudes.

\subsection{Equivariance Regularization}
\label{chap:equivariance}

L-GTA incorporates an equivariance regularization term to encourage consistency between transformations applied in latent and observation spaces. Specifically, applying a transformation in latent space followed by decoding should produce a result similar to decoding first and then applying the corresponding transformation in data space. Let $\mathcal{T}$ denote a perturbation drawn from the same family as the time series transformations (e.g., jitter, scaling), applied either in latent or observation space. We therefore enforce the approximate equivariance condition
\begin{equation}
\mathcal{D}(\mathcal{T}(v)) \approx \mathcal{T}(\mathcal{D}(v)),
\end{equation}
i.e., the decoder should be equivariant to these perturbations. During training, for each batch we sample a transformation $\mathcal{T}_{\eta}$ by drawing parameters $\eta$ from the predefined parameter space of the transformation family. We then compute the decoded observation $\hat{x} = \mathcal{D}(v)$ from the current latent mean $v$ and define the \emph{equivariance loss} as the mean squared error between decoding the perturbed latent and the perturbed decoding:
\begin{equation}
\mathcal{L}_{\mathrm{equiv}} = \mathbb{E}_{v,c,\mathcal{T}}\Bigl[ \bigl\| \mathcal{D}(\mathcal{T}(v)) - \mathcal{T}(\mathcal{D}(v)) \bigr\|^2 \Bigr].
\end{equation}
In practice, we use a single sampled $\mathcal{T}$ per batch and approximate the expectation by the average over the batch. The total training loss is
\begin{equation}
\mathcal{L} = \mathcal{L}_{\mathrm{recon}} + \beta \, \mathcal{L}_{\mathrm{KL}} + \lambda_{\mathrm{equiv}} \, \mathcal{L}_{\mathrm{equiv}},
\end{equation}
where $\mathcal{L}_{\mathrm{recon}}$ is the reconstruction loss, $\mathcal{L}_{\mathrm{KL}}$ is the KL divergence to the prior, and $\lambda_{\mathrm{equiv}} \geq 0$ is the equivariance weight. This term encourages the decoder to map latent perturbations to consistent changes in the reconstructed series, improving controllability when we apply the transformation functions.

\section{Experimental Setup}
\label{chap:results}

The experimental study is organized around the following research questions:
\begin{itemize}
    \item \textbf{Q1:} Do latent space manipulations in L-GTA produce synthetic series with controllable and interpretable transformation signatures?
    \item \textbf{Q2:} How does L-GTA compare with direct augmentation and SOTA augmentation methods in terms of downstream forecasting utility and computational efficiency?
    \item \textbf{Q3:} How does synthetic data generated by L-GTA compare with direct augmentation and SOTA augmentation methods in terms of fidelity and diversity?
    \item \textbf{Q4:} Which components of L-GTA contribute most to controllability and reconstruction quality?
\end{itemize}

\subsection{Datasets and Benchmark Methods}

We evaluate L-GTA and baselines on five time series datasets with different frequencies and scales. Tourism~\cite{athanasopoulos2011tourism} is the Australian Tourism dataset, comprising quarterly visitor numbers. Wiki2~\cite{pmlr-v139-rangapuram21a} consists of daily Wikipedia page-views. Labour~\cite{pmlr-v139-rangapuram21a} contains Australian monthly labour series. M3~\cite{makridakis2000m3} provides competition series from different domains in yearly and monthly frequencies. 

We compare L-GTA to the following. {Direct transformation} applies the same augmentations (jitter, scaling, magnitude warp, drift, trend) in observation space with fixed strength $\sigma$. It serves as an upper bound on controllability and transformation fidelity. {TimeGAN}~\cite{timegan}, {TimeVAE}~\cite{desai2021timevaevariationalautoencodermultivariate}, and {Diffusion-TS}~\cite{yuan2024diffusiontsinterpretablediffusiongeneral} represent three major families of generative time series models. TimeGAN is an adversarial model that learns a joint embedding space optimized through supervised and adversarial objectives to capture temporal dynamics. TimeVAE is a latent-variable model that generates sequences by sampling from a probabilistic representation learned with a variational autoencoder. Diffusion-TS is a recent diffusion-based approach that synthesizes time series through an iterative denoising process using a transformer encoder–decoder with disentangled temporal representations. We include these models as representative baselines for GAN-, VAE-, and diffusion-based generation, which constitute the dominant paradigms for synthetic time series generation.

We apply five commonly used time series augmentation transformations~\cite{Iwana_2021}. Specifically, we use jittering, scaling, magnitude warping, drift, and additive linear trend. Jittering adds Gaussian noise independently to each timestep. Scaling multiplies the entire series by a random scalar factor, modifying its amplitude. Magnitude warping introduces smooth nonlinear amplitude variations using a cubic spline defined by randomly sampled knots. Drift adds a cumulative Gaussian random walk, producing temporally correlated shifts in the signal level. Finally, additive linear trend introduces a random linear ramp over time, generating gradual increases or decreases in the series while keeping the trend centered to avoid systematic level shifts.

The component ablation evaluates eight L-GTA variants to isolate the effect of latent structure, equivariance, encoder design, dynamic features, and channel attention. Variants differ in whether they use a global or temporal latent representation, include equivariance regularisation, incorporate dynamic (time-varying) features, or apply channel-wise attention in the encoder. In particular, {Global + Equiv} uses a pooled latent representation with equivariant decoder training, while {L-GTA w/ dyn} combines temporal latent representations, equivariance, and dynamic features. Additional variants remove equivariance, simplify the encoder, or introduce channel attention to assess their impact on controllability and reconstruction. All variants share the same five transformations, $\sigma$ parameters, and evaluation protocol.

\subsection{Evaluation Protocols and Metrics}

\paragraph{Downstream Task.}
Our primary benchmark is a downstream forecasting task. For each dataset, the series are partitioned into windows, and we reserve a number of them for evaluation purposes. Forecasting accuracy is measured with Mean Absolute Scaled Error (MASE) using an LSTM forecaster trained over multiple random restarts. We consider two evaluation protocols. In the {Train-on-Synthetic, Test-on-Real (TSTR)} setting, the forecaster is trained exclusively on synthetic windows and evaluated on held-out real windows. In the {augmentation} setting, the forecaster is trained on the union of original and synthetic windows and is again evaluated on held-out real windows.

\paragraph{Synthesis Quality Metrics.}
Downstream forecasting alone does not fully characterize the quality of synthetic data. We therefore complement it with a feature-based quality analysis following prior work on synthetic time series evaluation~\cite{FACOCO2025102256}. Each time series is embedded using hand-crafted time series features extracted with TSFEL~\cite{barandas2020tsfel}, which we use to compare the synthetic and real datasets.

Fidelity metrics evaluate how closely the synthetic data distribution matches the real data. To assess structural similarity, we compute the cosine similarity between the first principal components of the real and synthetic feature sets. Values close to 1 indicate that the synthetic data captures the dominant direction of variation in the real data. We further report three distributional distance metrics, Wasserstein distance, Fréchet distance, and maximum mean discrepancy (MMD), to quantify discrepancies between the real and synthetic feature distributions. Diversity is evaluated using coverage, which estimates how well the synthetic samples span the feature space of the real data. Higher values indicate that the generated series represent a broader portion of the real data manifold.

\paragraph{Ablation Studies.}
To quantify controllability, let $\alpha$ denote the magnitude of a latent transformation and $\beta$ the resulting transformation strength. {Mean $\rho$} is the Spearman correlation $\rho = \mathrm{corr}_{\mathrm{rank}}(\alpha,\beta)$ computed over all evaluated transformations. {Monotonic \%} reports the proportion of transformation sequences for which the mapping $\alpha \mapsto \beta$ is monotonic. {Reconstruction MSE} is defined as $\frac{1}{T}\sum_{t=1}^{T}(z_t - \hat{z}_t)^2$.

Mean fingerprint distance measures how closely the decoded transformation reproduces the corresponding direct transformation based on the structure of the residuals. Let $r_t = \tilde{z}_t - z_t$ denote the residual between the transformed and original series. For each transformation type, we compute a three-dimensional residual fingerprint $\mathbf{f} = (f_1,f_2,f_3)$ that summarizes key structural properties of the residual sequence $r_t$. The first component, $f_1 = \mathrm{corr}(r_t,r_{t-1})$, is the lag-1 autocorrelation and captures the temporal smoothness of the residuals. The second component, $f_2$, is the coefficient of determination $R^2$ obtained from the regression $r_t = a + bt + \varepsilon_t$, which measures the presence of a linear temporal trend. The third component, $f_3 = \mathrm{corr}(|r_t|,|z_t|)$, quantifies amplitude dependence by assessing whether the magnitude of the residuals scales with the magnitude of the original signal. The fingerprint distance is defined as $d_{\mathrm{FP}} = \|\mathbf{f}^{\mathrm{model}} - \mathbf{f}^{\mathrm{direct}}\|_2$, and mean fingerprint distance reports the average value of $d_{\mathrm{FP}}$ across all evaluated transformations.

\section{Results and Discussion}

\paragraph{Controllability of latent transformations (Q1).}

\begin{figure*}[t]
\centering
\includegraphics[width=\textwidth]{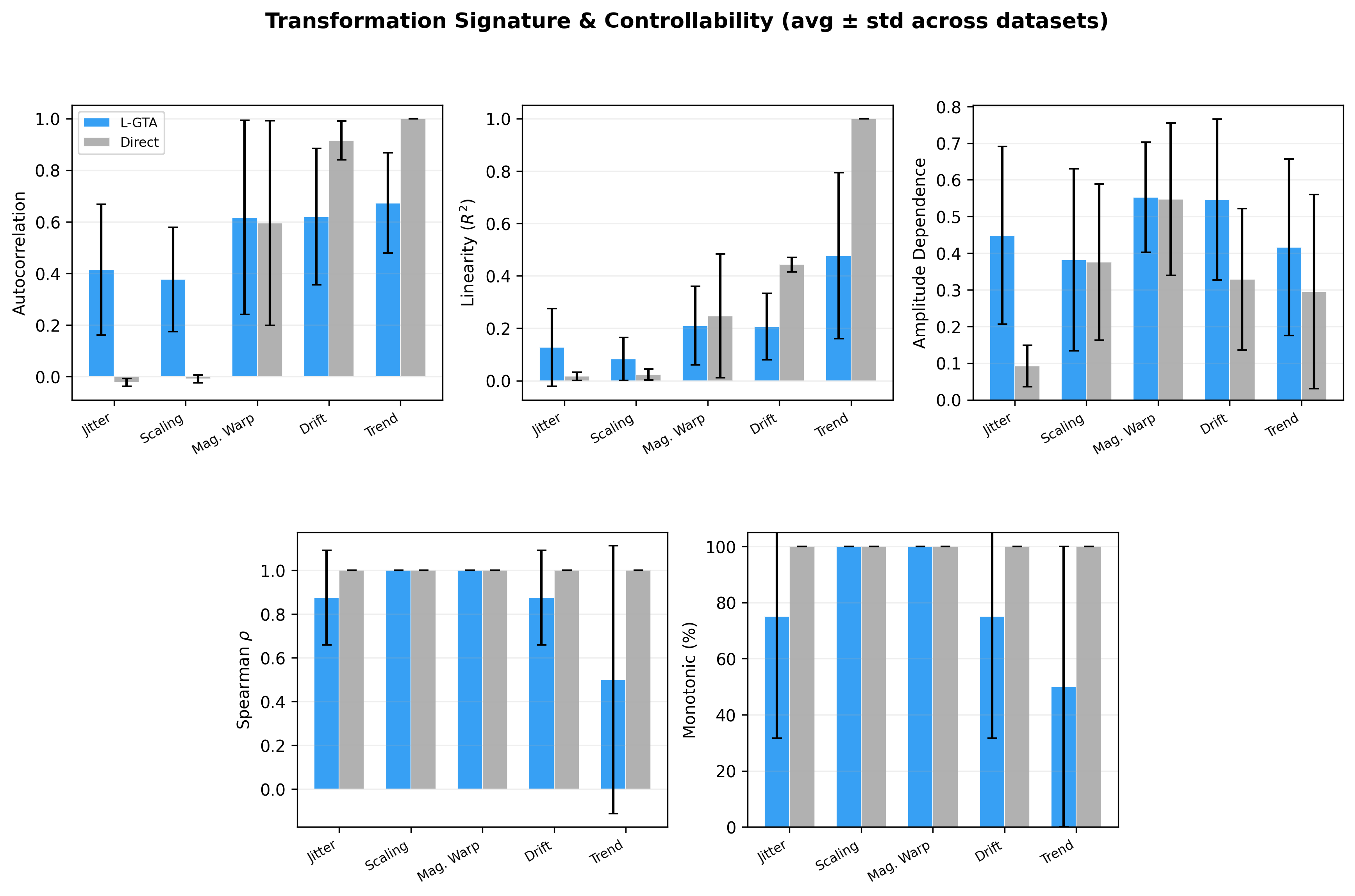}
\caption{
The top row shows residual fingerprint statistics comparing L-GTA (blue) with direct transformations (gray): lag-1 autocorrelation, temporal linearity ($R^2$), and amplitude dependence. The bottom row reports controllability metrics, including the Spearman correlation between transformation strength $\sigma$ and the resulting distributional shift, and the percentage of monotonic responses. 
}
\label{fig:transformation_fingerprint}
\end{figure*}

Figure~\ref{fig:transformation_fingerprint} evaluates whether latent manipulations in L-GTA reproduce the intended augmentation behavior. The residual fingerprints show that the model largely preserves the characteristic signatures of the transformations across datasets. Stochastic perturbations (e.g., jitter and scaling) remain weakly structured, while smooth transformations (e.g., magnitude warp, drift, and trend) produce strongly autocorrelated residuals. Similarly, transformations that introduce deterministic temporal structure, such as trend, remain identifiable through high linearity scores. Multiplicative effects (e.g., scaling and magnitude warp) also retain stronger dependence on signal amplitude. We also observe that the decoder introduces a regularization effect, attenuating strong smoothing effects while slightly increasing amplitude dependence for additive perturbations, most notably jitter.

The controllability metrics show that increasing the latent transformation strength leads to predictable changes in the generated series. For most transformations, the correlation between augmentation intensity and the resulting distributional shift remains high and the majority of datasets exhibit a monotonic response to the control parameter. This indicates that the latent space provides a reliable mechanism for modulating transformation strength.

\paragraph{Downstream forecasting utility and comparison to SOTA (Q2).}

We next evaluate how well synthetic series generated by L-GTA support downstream forecasting in both TSTR and augmentation regimes. Table~\ref{tab:downstream-summary} summarises aggregated results for all datasets. L-GTA attains the lowest combined MASE (2.99), among all methods, and is the only approach that ranks first in either TSTR or augmentation in every case. In contrast, strong generative baselines such as TimeVAE, TimeGAN, and Diffusion-TS, as well as the direct transformation baseline, exhibit higher mean errors, indicating that their synthetic data either underfit or over-regularise the forecasting task. These results show that latent-space augmentation with L-GTA contributes to improve downstream predictive performance relative to using only real data.

L-GTA also has a very low computational cost. SOTA generative methods are orders of magnitude slower: TimeVAE is about $8\times$ slower, Diffusion-TS over $10^3\times$ slower, and TimeGAN more than $5\times10^3\times$ slower.

\begin{table}[bt]
\centering
\caption{Aggregated downstream forecasting results. Mean and standard deviation of combined MASE across TSTR and augmentation. The next two columns report how many dataset–frequency cells each method achieved the best result in TSTR and in downstream forecasting. We report computational time relative to L-GTA (baseline = $0\times$).}
\label{tab:downstream-summary}
\begin{tabular}{@{}lcccc@{}}
\toprule
\textbf{Method} & \textbf{MASE (mean $\pm$ std)} & \textbf{Best (TSTR)} & \textbf{Best (Down.)} & \textbf{Comp. Time} \\
\midrule
L-GTA & $2.99 \pm 1.35$ & $3/5$ & $4/5$ & $0\times$ \\
TimeVAE & $4.06 \pm 1.88$ & $1/5$ & $0/5$ & $8.2\times$ \\
Original & $4.12 \pm 1.83$ & $0/5$ & $0/5$ & -- \\
Direct & $4.44 \pm 2.25$ & $1/5$ & $0/5$ & -- \\
Diffusion-TS & $5.33 \pm 2.50$ & $0/5$ & $1/5$ & $1085\times$ \\
TimeGAN & $5.98 \pm 3.35$ & $0/5$ & $0/5$ & $5487\times$ \\
\bottomrule
\end{tabular}
\end{table}

\paragraph{Synthesis quality and practical cost (Q3).}

L-GTA achieves the best balance between fidelity and diversity among the evaluated methods. It attains near-perfect structural similarity and the lowest distributional distances while maintaining the highest coverage of the real data manifold. In contrast, direct transformations provide moderate coverage but substantially weaker fidelity. Generative baselines such as TimeVAE and Diffusion-TS achieve good fidelity but cover a smaller portion of the real distribution, whereas TimeGAN attains high coverage at the cost of large distributional mismatches. Hence, L-GTA combines the strengths of both approaches, producing synthetic data that is both realistic and diverse.

\begin{table}[tb]
\centering
\caption{Synthesis-quality metrics averaged over all datasets. Coverage and cosine similarity indicate diversity and structural similarity (higher is better), while Wasserstein, Fréchet, and MMD measure distributional distance (lower is better).}
\label{tab:quality-summary}
\begin{tabular}{lccccc}
\toprule
\textbf{Method} & \textbf{Coverage} & \textbf{Cos.\ sim.} & \textbf{Wasserstein} & \textbf{Fr\'echet} & \textbf{MMD} \\
\midrule
L-GTA        & 0.97 & 1.00 & 0.01 & 0.01 & 0.02 \\
Direct       & 0.84 & 0.98 & 0.52 & 0.51 & 0.57 \\
TimeVAE      & 0.83 & 0.99 & 0.01 & 0.01 & 0.03 \\
Diffusion-TS & 0.75 & 1.00 & 0.01 & 0.01 & 0.04 \\
TimeGAN      & 0.95 & 0.99 & 0.56 & 0.54 & 0.56 \\
\bottomrule
\end{tabular}
\end{table}

\paragraph{Contribution of architectural components (Q4).}

The ablation study highlights which architectural components are most critical for controllability and reconstruction. As shown in Table~\ref{tab:ablation-summary}, the L-GTA baseline already achieves strong controllability with low fingerprint distance. Removing equivariance or using simpler temporal encoders substantially degrades performance, leading to higher reconstruction errors (122--166\%) and weaker controllability. Dynamic conditioning maintains high controllability but increases reconstruction error, suggesting that its main benefit is aligning transformations with time-varying covariates rather than improving reconstruction fidelity. Finally, adding channel-wise attention provides no clear advantage and can reduce controllability when dynamic conditioning is removed. These findings further support our design choice of relying on temporal attention only for stable augmentation behavior.

\begin{table}[h]
\centering
\caption{Component ablation summary. Metrics are averaged over all (dataset, frequency) pairs where each variant appears. Reconstruction MSE is reported relative to the L-GTA model (100\%). FP dist = fingerprint distance to the corresponding direct transformation.}
\label{tab:ablation-summary}
\resizebox{\textwidth}{!}{%
\begin{tabular}{@{}lcccc@{}}
\toprule
\textbf{Description} & \textbf{Mean $\rho$} & \textbf{Monotonic \%} & \textbf{Recon.\ MSE} & \textbf{Mean FP dist.} \\
\midrule
L-GTA & 0.85 & 80.0 & 100\% & 0.49 \\
L-GTA, w/ dyn & 0.95 & 90.0 & 147\% & 0.49 \\
Global + Equiv & 0.95 & 90.0 & 96\% & 0.61 \\
Global, No Equiv & 0.78 & 80.0 & 122\% & 0.88 \\
Temporal, No Equiv & 0.55 & 55.0 & 128\% & 0.85 \\
Simple Enc (temporal) & 0.70 & 85.0 & 166\% & 0.83 \\
L-GTA + chan attn, w/ dyn & 0.82 & 80.0 & 102\% & 0.90 \\
L-GTA + chan attn & 0.45 & 65.0 & 221\% & 0.65 \\
\bottomrule
\end{tabular}%
}
\end{table}

\section{Conclusions}
\label{chap:concl}

We introduced {L-GTA}, a method that generates controllable semi-synthetic time series through latent transformations. L-GTA combines temporal self-attention, Bi-LSTMs, and a variational autoencoder with an equivariance objective. It allows time series transformations to be applied in its latent space and decoded into realistic sequences that preserve the statistical and temporal structure of the original data.

Our results show that latent manipulations in L-GTA produce consistent and interpretable transformation effects. The residual fingerprint analysis demonstrates that the model preserves the structural signatures of the intended augmentations across datasets. L-GTA also generates synthetic data that is both realistic and diverse, achieving the best balance between fidelity and coverage among the evaluated methods. In addition, the generated data proves effective for downstream tasks: L-GTA consistently achieves the lowest forecasting error while remaining computationally efficient compared to alternative generative approaches. The ablation study further confirms that equivariance regularization and structured temporal encoding are key components for achieving stable and controllable latent transformations.

Several directions remain for future work. One avenue is to extend L-GTA to support a broader class of transformations, including domain-specific transformations relevant to particular applications. Another is to evaluate L-GTA across a wider range of domains and downstream tasks, such as anomaly detection or classification. This would provide deeper insight into the robustness and generalization of latent transformation-based data augmentation.

\begin{credits}
\subsubsection{\ackname} This work is funded by national funds through FCT – Foundation for Science and Technology, I.P., within the scope of the research unit UID/00326 - Centre for Informatics and Systems of the University of Coimbra, \url{https://doi.org/10.54499/UID/00326/2025}. This work was also funded by the Agenda “Center for Responsible AI”, nr. C645008882-00000055, investment project nr. 62, financed by the Recovery and Resilience Plan (PRR) and by European Union - NextGeneration EU. Funded by the European Union – NextGenerationEU. Views and opinions expressed are however those of the author(s) only and do not necessarily reflect those of the European Union or the European Commission. Neither the European Union nor the European Commission can be held responsible for them. 
\end{credits}

\end{document}